\documentclass[parskip=full]{scrartcl}

\pdfoutput=1

\title{Geometric SMOTE: Effective oversampling for imbalanced learning through a geometric extension of SMOTE}

\author{
	Georgios Douzas\(^{1}\), Fernando Bacao\(^{1*}\)
	\\
	\small{\(^{1}\)NOVA Information Management School, Universidade Nova de Lisboa}
	\\
	\small{*Corresponding Author}
	\\
	\\
	\small{Postal Address: NOVA Information Management School, Campus de Campolide, 1070-312 Lisboa, Portugal}
	\\
	\small{Telephone: +351 21 382 8610}
}

\usepackage{breakcites}
\usepackage{float}
\usepackage{graphicx}
\usepackage{geometry}
\usepackage{amsmath}
\newcommand{\inlineeqnum}{\refstepcounter{equation}~~\mbox{(\theequation)}}
\usepackage{enumitem}
\newlist{algorithm}{enumerate}{10}
\setlist[algorithm]{label*=\arabic*.}
\usepackage{booktabs}
\usepackage{pgfplotstable}
\usepackage{longtable}
\usepackage{tabu}
\usepackage{bibentry}
\date{}

\begin{document}

\maketitle

\begin{abstract}
Classification of imbalanced datasets is a challenging task for standard algorithms. Although many methods exist to address this problem in different ways, generating artificial data for the minority class is a more general approach compared to algorithmic modifications. SMOTE algorithm and its variations generate synthetic samples along a line segment that joins minority class instances. In this paper we propose Geometric SMOTE (G-SMOTE) as a generalization of the SMOTE data generation mechanism. G-SMOTE generates synthetic samples in a geometric region of the input space, around each selected minority instance. While in the basic configuration this region is a hyper-sphere, G-SMOTE allows its deformation to a hyper-spheroid and finally to a line segment, emulating, in the last case, the SMOTE mechanism. The performance of  G-SMOTE is compared against multiple standard oversampling algorithms. We present empirical results that show a significant improvement in the quality of the generated data when G-SMOTE is used as an oversampling algorithm.
\end{abstract}

\section{Introduction}
Learning from imbalanced data is a non trivial and important problem for the research community and the industry practitioners \cite{Chawla2003}. An imbalanced learning problem is defined as a classification task for binary or multi-class datasets where a significant asymmetry exists between the number of instances for the various classes. The dominant class is called the majority class while the rest of the classes are called the minority classes \cite{Chawla2003}. The Imbalance Ratio (IR), defined as the ratio between the majority class and each of the minority classes, depends on the type of application and for binary problems values between 100 and 100.000 have been observed \cite{Chawla2002}, \cite{Barua2014}.

Multiple real-world applications are connected to imbalanced data. Examples include medical diagnosis, information retrieval systems, fraud detection, detection of oil spills in radar images, direct marketing, automatic classification of land use and land cover in remote sensing images, detection of rare particles in experimental high-energy physics, telecommunications management and bioinformatics \cite{Akbani2004}, \cite{He2009}, \cite{Clearwater1991}, \cite{Graves2016}, \cite{Verbeke2012}, \cite{Zhao2008}. Standard learning methods induce a bias in favor of the majority class. Specifically, during the training phase, the minority classes contribute less to the minimization of the objective function. Also the distinction between noisy and minority class instances is often difficult. The result is that the performance of the classifiers, evaluated on metrics appropriate for imbalanced data, is low. An important observation is that in many of these applications the misclassification cost of the minority classes is often higher than the misclassification cost of the majority class \cite{Domingos1999}, \cite{Ting2002}. Therefore the methods that address the class imbalance problem aim to increase the classifier's accuracy for the minority classes.

There are three main approaches to deal with the class imbalance problem \cite{Fernandez2013}. The first is the modification or creation of algorithms that reinforce the learning towards the minority class. The second approach is the application of cost-sensitive methods at the data or algorithmic level in order to minimize higher cost errors. The third and more general approach is the modification at the data level by re-balancing the class distribution through under-sampling, over-sampling or hybrid methods.

Our focus in this paper is oversampling techniques, which result in the generation of artificial data for the minority class. Standard oversampling methods, inspired by Synthetic Minority Oversampling Technique (SMOTE) algorithm \cite{Chawla2002}, generate synthetic samples along the line segment that joins minority class samples. Contrary to this, the approach proposed in this paper, G-SMOTE, selects a minority class instance, as well either a majority or a minority sample, and defines a flexible geometric region around it. The flexibility of the geometric boundaries is controlled by an appropriate parametrization of the algorithm.

For the evaluation of G-SMOTE as an oversampling method an experimental analysis is performed, based on various publicly available datasets from Machine Learning Repository. Then the proposed method is compared to SMOTE algorithm, Borderline SMOTE \cite{Han2005} and ADASYN \cite{He2008}. For the classification of the binary class data two classifiers and three evaluation metrics are applied.

The sections in the paper are organized as follows. In section 2, an overview of related previous works and existing sampling methods is given. In Section 3, the motivation for G-SMOTE is presented, while section 4 describes the proposed method in detail. In section 5 the experimental results are presented and conclusions from their analysis are provided in section 6.

\section{Related work}
In this section we provide a short review of the oversampling methods. A review of the other methods can be found in \cite{Galar2012}, \cite{Chawla2005}. Oversampling methods generate synthetic examples for the minority class and add them to the training set. The simplest approach, Random Oversampling, duplicates randomly selected minority class instances. The disadvantage of this approach is that the exact replication of training examples can increase the risk of overfitting since the classifier is exposed to the same information. 

An alternative approach that aims to eliminate this problem and generate new data is SMOTE. Synthetic data are generated along the line segment that joins minority class samples. SMOTE has the disadvantage that, since the separation between majority and minority class clusters is not often clear, noisy samples may be generated \cite{He2009}. To avoid this scenario various modifications of SMOTE have been proposed. SMOTE + Edited Nearest Neighbor \cite{Batista2004} combination applies the edited nearest neighbor rule \cite{Wilson1972} after the generation of artificial examples through SMOTE to remove any misclassified instances, based on the classification by its three nearest neighbors. Safe-Level SMOTE \cite{Bunkhumpornpat2009} modifies the SMOTE algorithm by applying a weight degree, the safe level, in the data generation process. Borderline-SMOTE \cite{Han2005}, MWMOTE (Majority Weighted Minority Oversampling Technique for Imbalanced Data Set Learning) \cite{Barua2014}, ADASYN and its variation KernelADASYN \cite{Tang2015} aim to avoid the generation of noisy samples by identifying the borderline instances of the majority and minority classes that in turn are used to identify the informative minority class samples. 

The methods above address the problem of between-class imbalance \cite{Nekooeimehr2016}. Another type of problem is the within-class imbalance  \cite{Nekooeimehr2016}, \cite{Bunkhumpornpat2012}, \cite{Cieslak2008}, \cite{Jo2004} i.e. when sparse or dense sub-clusters of minority or majority instances exist.  Clustering based oversampling methods that deal with the between-class imbalance problem have recently been proposed. These methods are initially partitioning the input space and then apply sampling methods in order to adjust the size of the various clusters. Cluster-SMOTE \cite{Cieslak2006} applies the k-means algorithm and then generates artificial data by applying SMOTE in the clusters. Similarly DBSMOTE \cite{Bunkhumpornpat2012} uses the DB-SCAN algorithm to discover arbitrarily shaped clusters and generates synthetic instances along a shortest path from each minority class instance to a pseudo-centroid of the cluster. A-SUWO \cite{Nekooeimehr2016} creates clusters of the minority class instances with a size, which is determined using cross validation and generates synthetic instances based on a proposed weighting system. SOMO \cite{Douzas2017} creates a two dimensional representation of the input space and based on it, applies the SMOTE procedure to generate intra-cluster and inter-cluster synthetic data that preserve the underlying manifold structure. 

Other types of oversampling approaches are based on ensemble methods \cite{Wang2015}, \cite{Sun2015} such as SMOTEBoost \cite{Chawla2003} and DataBoost-IM \cite{Guo2004}. Finally a different type of oversampler is based on the Conditional Generative Adversarial Networks \cite{Douzas2017}. Contrary to the previous methods, this method aims to approximate the true data distribution and generate data for the minority class.

\section{Motivation}
In the previous section various informative oversampling methods were presented as effective way to re-balance the data distribution. However, there are scenarios where SMOTE and its variations may encounter a variety of problems. This section describe some of these cases and motivates the proposed G-SMOTE algorithm. Some of the inefficiencies of the SMOTE based algorithms are the following:  

1. \textit{Generation of noisy examples that penetrate in the area of the majority class examples.}

SMOTE, as well as many other methods, applies the \(k\)-nearest neighbor approach during the sample generation phase. In order to generate an artificial example \(x_{gen}\) from an existing minority class example \(x\), first the kNN approach randomly selects another minority class sample \(x'\) from the \(k\)-nearest neighbors of \(x\). Then \(x_{gen}\) is generated by using a linear interpolation of \(x\) and \(x'\) which can be expressed as \( x_{gen} = x + \alpha \cdot (x' - x) \), where \(a\) is a random number drawn from the uniform distribution in the \([0, 1]\) interval. Thus \( x_{g} \) lies in the line segment between \(x\) and \(x'\). Nevertheless, the appropriate value of \(k\) cannot be determined in advance and the results of oversampling are sensitive to it, as is shown in the next examples. A large value can result in the generation of noisy samples since \(x'\) might be a noisy sample that has penetrated in the majority class area as is shown in Fig. 1.  Even a small value of \(k\) does not always avoid the above scenario since x might already be a noisy sample. This scenario can also be seen in Fig. 2. The generation of noisy samples can reduce the accuracy of predictions for both the minority and majority classes \cite{Kotsiantis2006}.

\begin{figure}[H]
	\centering
	\includegraphics[width=12cm, keepaspectratio]{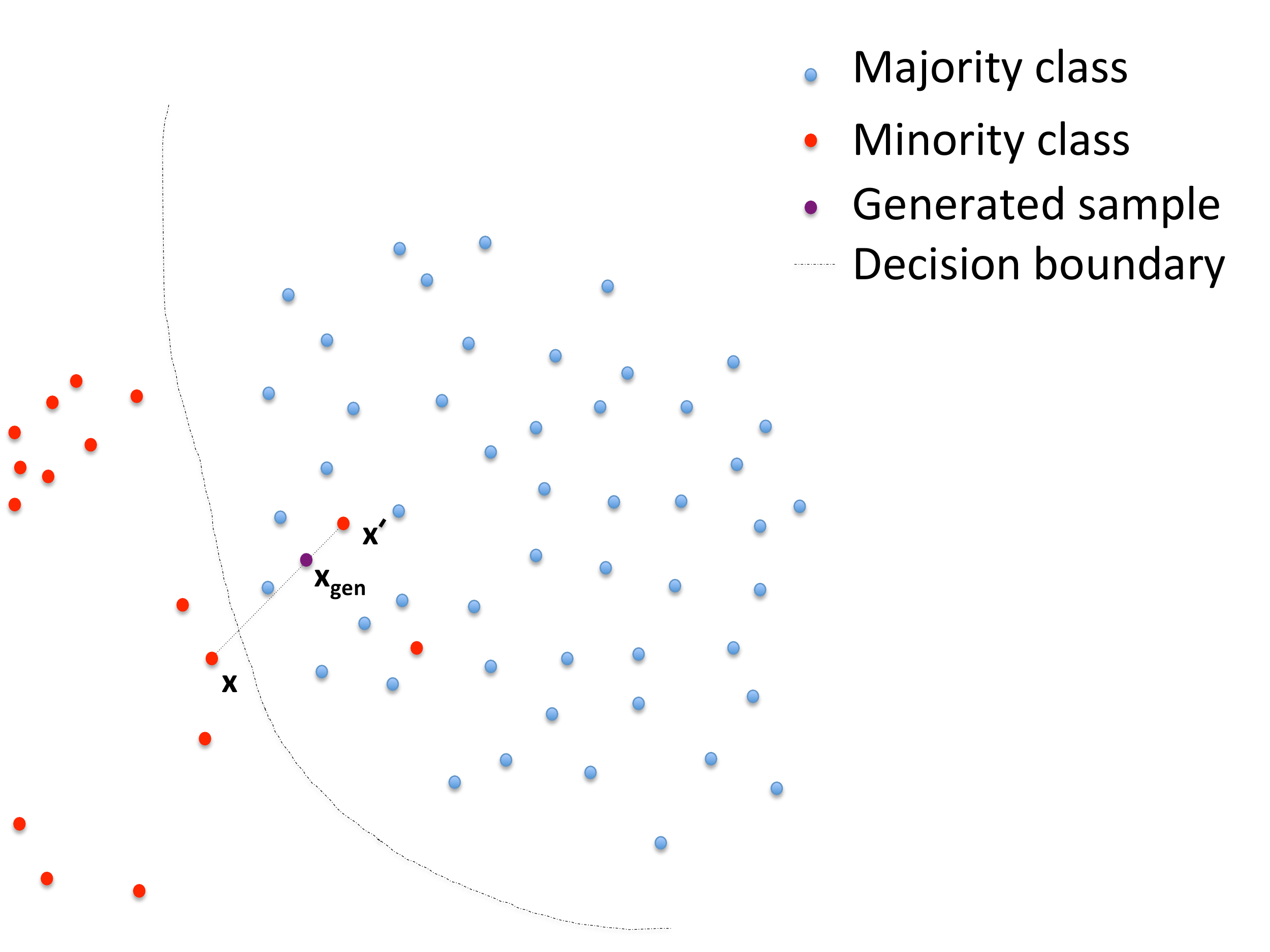}
	\captionbelow{An instance near the decision boundary and one of its 4-nearest neighbors are selected randomly. A noisy observation is generated.}
\end{figure}

\begin{figure}[H]
	\centering
	\includegraphics[width=12cm, keepaspectratio]{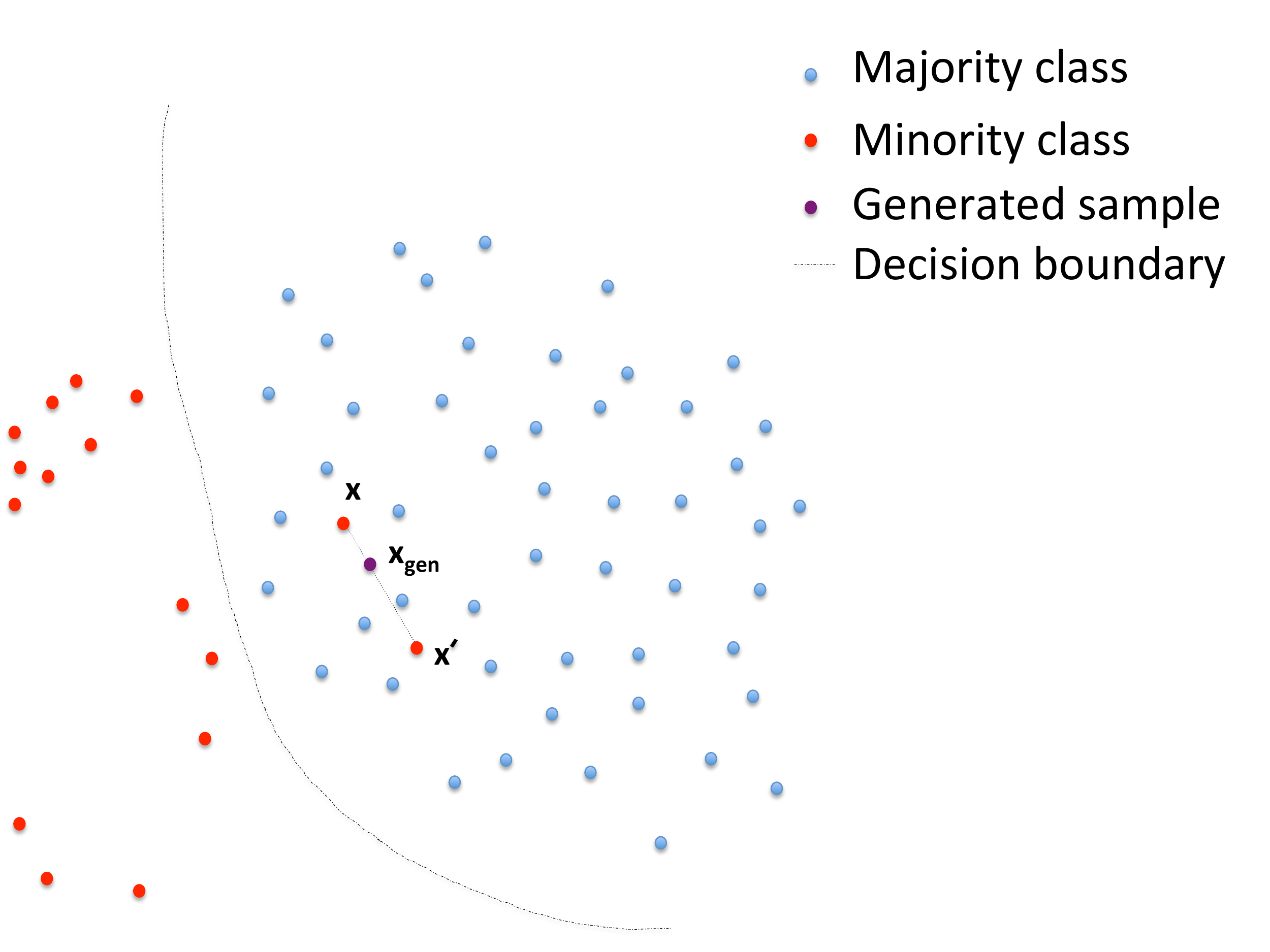}
	\captionbelow{A 2-nearest neighbor does not avoid the generation of noise when noisy samples are initially selected.}
\end{figure}

2. \textit{Generation of nearly duplicated samples.}

As was mentioned above a small value of \(k\) can be chosen in order to avoid the generation of noisy samples. Another approach is to generate a new instance between a selected observation and one of its neighbors from the same cluster \cite{Barua2014}. Both of these choices have the consequence that the generation of synthetic examples might occur in dense minority class areas as is shown in Fig. 3. These samples are less useful because they do not add any new information to the imbalanced data set and may lead to overfitting. Thus it is desirable to expand the data generation process in areas where minority examples are absent. Fig. 4 presents a scenario where the number of \(k\)-nearest neighbors is increased in order to generate two inter-cluster instances. This attempt results in the generation of a noisy instance.

\begin{figure}[H]
	\centering
	\includegraphics[width=12cm, keepaspectratio]{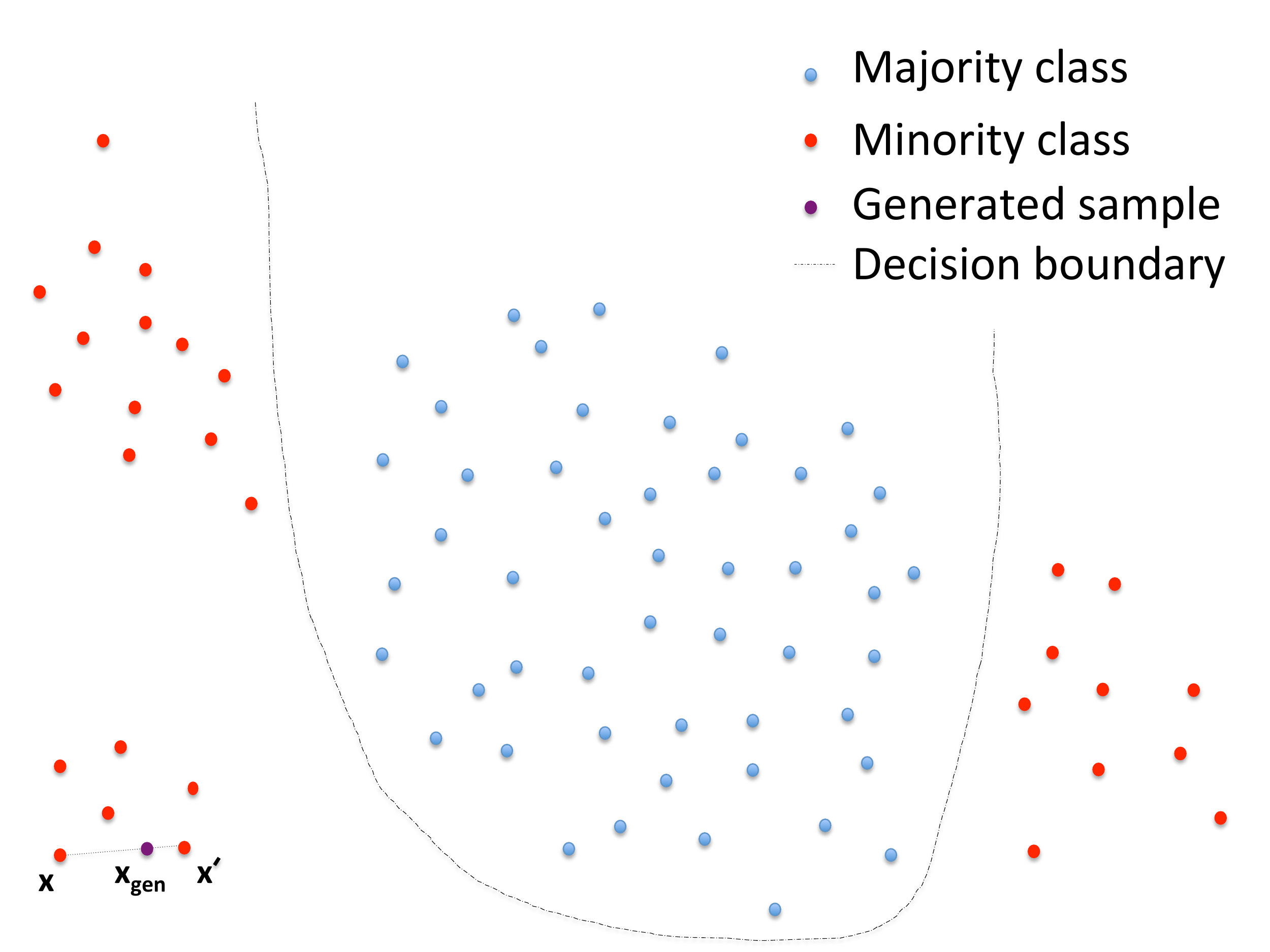}
	\captionbelow{An instance belonging to a minority class cluster and one of its 5-nearest neighbors are selected. An observation belonging to the same cluster is generated.}
\end{figure}

\begin{figure}[H]
	\centering
	\includegraphics[width=12cm, keepaspectratio]{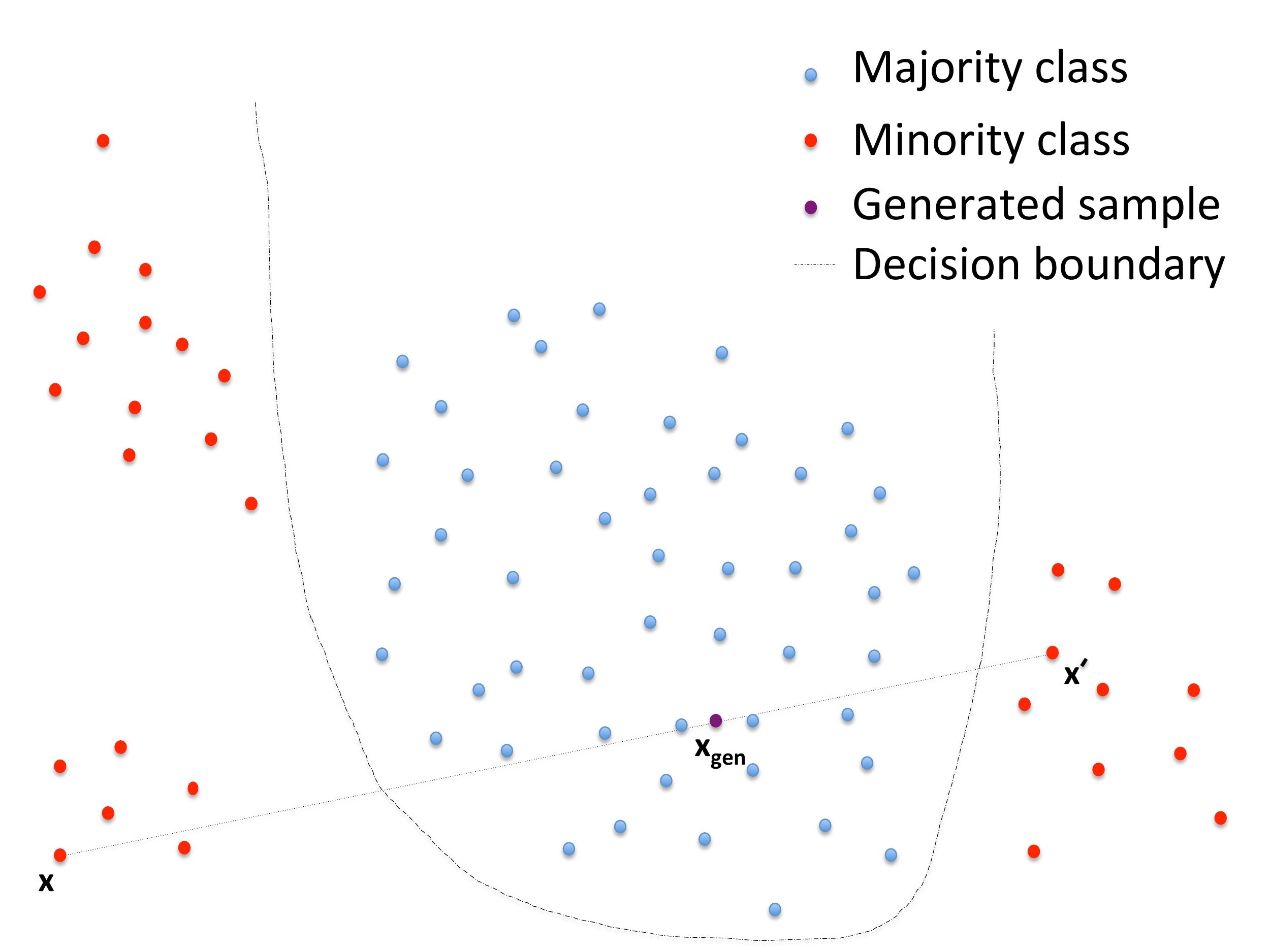}
	\captionbelow{An attempt to generate intercluster instances by increasing the number of k-nearest neighbors of the selected instances. One of the generated instances penetrates in the majority class area.}
\end{figure}

\section{The proposed method}

In the previous section some insufficiencies of the existing methods that may apply in various cases were described. We propose a novel data generation procedure, G-SMOTE, which is an extension of the SMOTE algorithm and has three main objectives:

1. \textit{To define a safe area around each selected minority class instance such that the generated artificial minority instances inside this area are not noisy.}

2. \textit{To increase the variety of generated samples by expanding the minority class area.}

3. \textit{To parametrize the above characteristics based on a small number of transformations with a geometrical interpretation.}

\subsection{G-SMOTE algorithm}

The complete algorithm in pseudo-code is the following:

\textbf{Algorithm}

\( G-SMOTE \Big( S_{maj}, S_{min}, N, k, \alpha_{trunc}, \alpha_{def}, \alpha_{sel} \Big) \)

\( \textbf{Inputs} \)

\begin{algorithm}
	
	\item \( S_{maj} \): Set of majority class samples.
	
	\item \( S_{min} \): Set of minority class samples.
	
	\item \( N \): Total number of synthetic samples to be generated.
	
	\item \( k \): Number of nearest neighbors.
	
	\item \( \alpha_{trunc} \): Truncation factor with \( -1 \leq \alpha_{trunc} \leq 1 \).
	
	\item \( \alpha_{def} \): Deformation factor with \( 0 \leq \alpha_{def} \leq 1 \) .
	
	\item \( \alpha_{sel} \): Neighbor selection strategy with \( \alpha_{sel} \in \Big\{ minority, majority, combined \Big\} \).
	
\end{algorithm}

\textbf{Procedure begin}

\begin{algorithm}

	\item Shuffle \( S_{min} \)

	\item Repeat until \( N \) minority instances are selected, each multiple times if necessary, in the order that appear in \( S_{min} \):

	\begin{algorithm}

		\item Let \( \textbf{x}_{center} \in \ S_{min} \) the selected minority class instance of \( p \) components.

		\item If \( \alpha_{sel} = minority \):
		
		\begin{algorithm}
			
			\item Calculate the \( k \) nearest neighbors of \( \textbf{x}_{center} \) from \( S_{min} \). Let \( \textbf{x}_{surface} \) one of them, randomly selected.
			
			\item Calculate a radius from the relation \( R \leftarrow d(\textbf{x}_{center}, \textbf{x}_{surface})  \).
			
		\end{algorithm}
	
		\item Else if \( \alpha_{sel} = majority \):
		
		\begin{algorithm}
			
			\item Calculate \( \textbf{x}_{surface} \) as the nearest neighbor of \( \textbf{x}_{center} \) from \( S_{maj}  \).
			
			\item Calculate a radius from the relation \( R \leftarrow d(\textbf{x}_{center}, \textbf{x}_{surface}) \).
			
		\end{algorithm}
	
		\item Else if \( \alpha_{sel} = combined \):
		
		\begin{algorithm}
			
			\item Calculate the \( k \) nearest neighbors of \( \textbf{x}_{center} \) from \( S_{min} \). \ Let \( \textbf{x}_{min} \) one of them, randomly selected.
			
			\item Calculate the euclidean distance \( d_{min} \leftarrow d(\textbf{x}_{center}, \textbf{x}_{min}) \).
			
			\item Calculate \( \textbf{x}_{maj} \) as the nearest neighbor of \( \textbf{x}_{center} \) from \( S_{maj} \).
			
			\item Calculate the euclidean distance \( d_{maj} \leftarrow d(\textbf{x}_{center}, \textbf{x}_{maj}) \).
			
			\item Calculate a radius from the relation \( R \leftarrow d(\textbf{x}_{center}, \textbf{x}_{surface}) \) where \\ \( \textbf{x}_{surface} \leftarrow \displaystyle argmin_{\textbf{x}_{min}, \textbf{x}_{maj}} \Big(d_{min}, d_{maj} \Big) \).
			
		\end{algorithm}
	
		\item Generate a synthetic sample \( \textbf{x}_{gen} \leftarrow hyperball_{(center = \textbf{0}, radius = 1)}() \).
		
		\item Transform\( ^{1} \) the synthetic sample by \( \textbf{x}_{gen} \leftarrow truncate(\textbf{x}_{gen}, \textbf{x}_{center}, \textbf{x}_{surface}, \alpha_{trunc}) \).
		
		\item Transform\( ^{2} \) the synthetic sample by \( \textbf{x}_{gen} \leftarrow deform(\textbf{x}_{gen}, \textbf{x}_{center}, \textbf{x}_{surface}, \alpha_{def}) \).
		
		\item Transform\( ^{3} \) the synthetic sample by \( \textbf{x}_{gen} \leftarrow translate(\textbf{x}_{gen}, \textbf{x}_{center}, R) \).
		
		\item Add the sample \( \textbf{x}_{gen} \) to the set of generated samples \( S_{gen} \).

	\end{algorithm}

\end{algorithm}

\textbf{Output}

The set \( S_{gen} \) of generated synthetic examples.

\textbf{Geometric functions}\( ^{1, 2, 3} \)  

Definitions for \( truncate \) and \( deform \) functions:

\begin{itemize}

	\item[] Define the unit vector \( \textbf{e}_{\scriptscriptstyle//} \leftarrow \frac{\textbf{x}_{surface} - \textbf{x}_{center}}{|\textbf{x}_{surface} - \textbf{x}_{center}|} \inlineeqnum\label{eqn:unit} \).

	\item[] Define the projection of \( \textbf{x}_{gen} \) to \( \textbf{e}_{\scriptscriptstyle//} \) as \( x_{\scriptscriptstyle//} = \textbf{x}_{gen} \cdot \textbf{e}_{\scriptscriptstyle//} \) where \( \cdot \) is the euclidean dot product.

	\item[] Define \( \textbf{x}_{\scriptscriptstyle//} \leftarrow x_{\scriptscriptstyle//} \textbf{e}_{\scriptscriptstyle//}  \inlineeqnum\label{eqn:parallel} \) and \( \textbf{x}_{\bot} \leftarrow \textbf{x}_{gen} - \textbf{x}_{\scriptscriptstyle//}  \) .

\end{itemize}

Function \( hyperball_{(center = \textbf{0}, radius = 1)}() \):

\begin{itemize}

	\item[] Generate a vector \( \textbf{v}_{normal} \leftarrow (v_{1}, \cdots, v_{p}) \) of \( p \) random numbers from the normal distribution \( N(0, 1) \).

	\item[] Calculate the unit vector \( \textbf{e}_{sphere} \leftarrow \frac{\textbf{v}_{normal}}{|\textbf{v}_{normal}|} \inlineeqnum\label{eqn:sphere} \) where \( |\cdot| \) is the euclidean norm.

	\item[] Calculate the vector \( \textbf{x}_{gen} \leftarrow r^{1 / p} \textbf{e}_{sphere} \inlineeqnum\label{eqn:ball} \) where \( r \) is a random number from the uniform distribution \( U(0, 1) \).

	\item[] Return \( \textbf{x}_{gen} \).

\end{itemize}

Function \( ^{1} \)\( truncate(\textbf{x}_{gen}, \textbf{x}_{center}, \textbf{x}_{surface}, \alpha_{trunc}) \):

\begin{itemize}

	\item[] If \( \textbf{x}_{surface} \neq \textbf{x}_{center} \):

	\begin{itemize}

		\item[] If \( |\alpha_{trunc} - x_{\scriptscriptstyle//}| > 1: \inlineeqnum\label{eqn:condition} \)

		\begin{itemize}

			\item[] Assign \( \textbf{x}_{gen} \leftarrow \textbf{x}_{gen} - 2 \textbf{x}_{\scriptscriptstyle//} \inlineeqnum\label{eqn:truncate} \).
			
		\end{itemize}

	\end{itemize}
	
	\item[] Return \( \textbf{x}_{gen} \) .
	
\end{itemize}

Function \( ^{2} \)\( deform(\textbf{x}_{gen}, \textbf{x}_{center}, \textbf{x}_{surface}, \alpha_{dist}) \):

\begin{itemize}

	\item[] If \( \textbf{x}_{surface} \neq \textbf{x}_{center}\):
		
	\begin{itemize}	
		
		\item[] Assign \( \textbf{x}_{gen} \leftarrow \textbf{x}_{gen} - \alpha_{def} \textbf{x}_{\bot} \inlineeqnum\label{eqn:deform} \).
		
	\end{itemize}

	\item[] Return \( \textbf{x}_{gen} \) . 
	
\end{itemize}

Function \( ^{3} \)\( translate(\textbf{x}_{gen}, \textbf{x}_{center}, R) \):

\begin{itemize}

\item[] Return \( \textbf{x}_{center} +  R \cdot \textbf{x}_{gen} \inlineeqnum\label{eqn:translate} \).

\end{itemize}

\subsection{Details and justification of the G-SMOTE algorithm}

As explained above, SMOTE compared to Random Oversampling, improves the diversity of generated samples by linearly interpolating generated samples between two minority class instances. However on high-dimensional data SMOTE does not change the class-specific mean values while it decreases the data variability and it introduces correlation between samples \cite{Blagus2013}. Contrary to this, G-SMOTE extends the linear interpolation mechanism by introducing a geometric region where the data generation process occurs. At the most general choice of hyper-parameters, this geometric region of the input space is a truncated hyper-spheroid. The various steps of the G-SMOTE algorithm can be described in detail as follows:

1, 2: The data are shuffled and the process described below is repeated \( N \)times until \( N \) artificial points have been generated.

2.1: A minority class instance \( \textbf{x}_{center} \) is selected as the center of a geometric region. The order of selection follows the order of the data set points after shuffling in step 1. Therefore if \( N \) is greater than \( S_{min} \), then some of the minority class samples will be selected more than once.

2.2: In this case the neighbor selection strategy is based only on the minority class and it is identical to the selection strategy of SMOTE. Initially the \( k \) nearest neighbors of \( \textbf{x}_{center} \) from the set \( S_{min} \) are identified. One of them, \( \textbf{x}_{surface} \), is randomly selected and its distance from \( \textbf{x}_{center} \) is defined as the radius \( R \). Fig. 5 presents an example of a minority class instance selection among the \( k = 4 \ \) nearest neighbors of \( \textbf{x}_{center} \). The time complexity of this selection strategy depends on the choice of the algorithm and increases with the dimensionality of the input space as well as the value of the \( k \) parameter \cite{Vaidya1989}. Therefore for a wide set of realistic cases restricting the search of nearest neighbors to the minority class has a lower computational cost than including the majority class instances in the search space.

\begin{figure}[H]
	\centering
	\includegraphics[width=12cm, keepaspectratio]{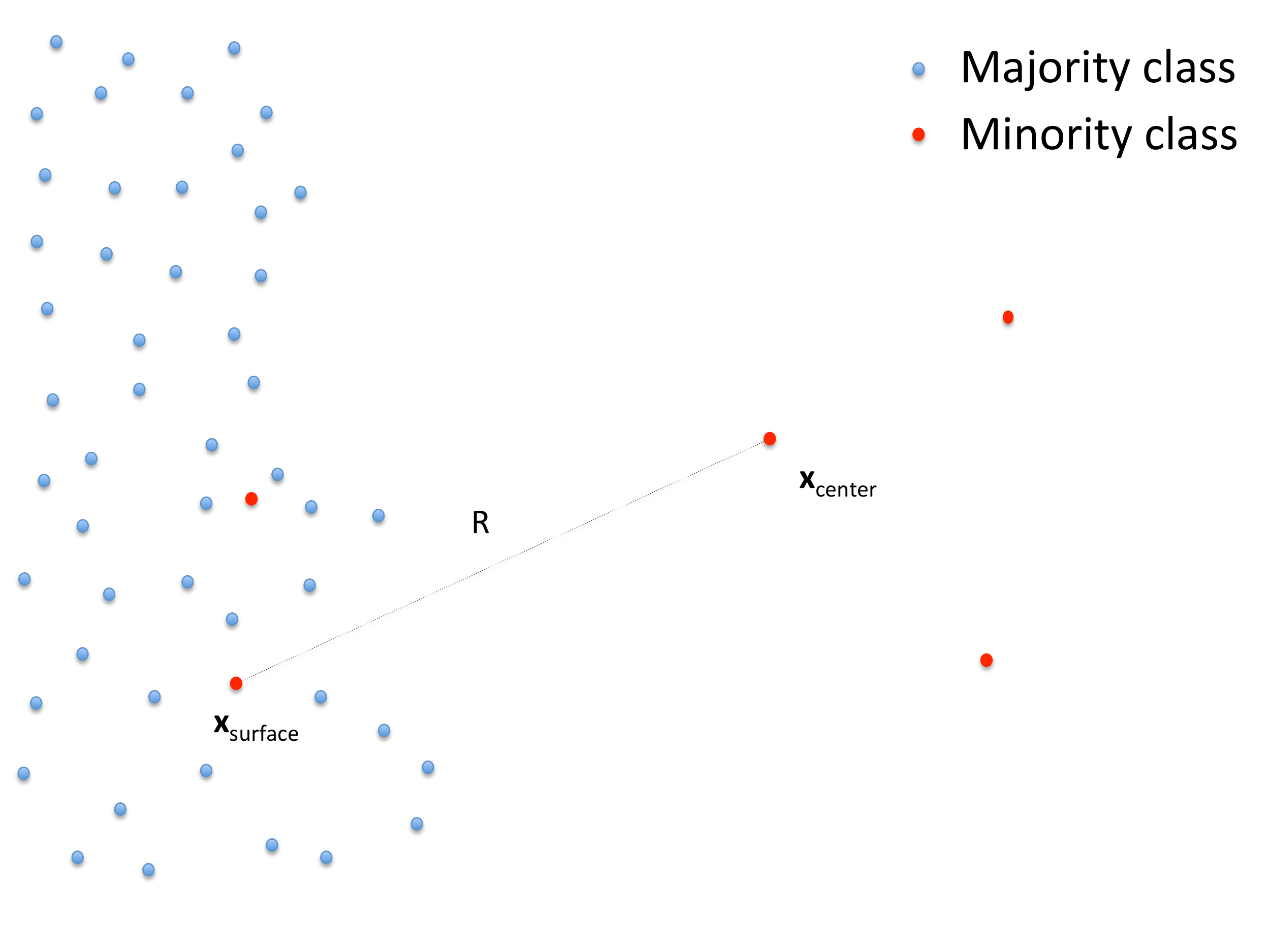}
	\captionbelow{An example of the minority selection strategy. A minority class instance is defined as the center of the hyper-spheroid and one of its k = 4 minority class nearest neighbors is selected as the surface point. Their distance is equal to the radius R of the hyper-spheroid.}
\end{figure}

2.3: As explained in section 3, one of the drawbacks of the minority selection strategy is that it may lead to the generation of data penetrating deeply in the majority class area. The majority selection strategy eliminates this scenario. More specifically, the nearest neighbour of \( \textbf{x}_{center} \) from the set \( S_{maj} \) is identified as \( \textbf{x}_{surface} \) and its distance from \( \textbf{x}_{center} \) is defined as the radius \( R \). The consequence of this selection is that when a random minority class point is generated inside this radius, it is ensured that its distance from \( \textbf{x}_{center} \) is not higher than the distance between \( \textbf{x}_{center} \) and any majority class instance. On the other hand, since any information about the minority class is discarded, this strategy might aggressively expand the minority class area, resulting effectively to noise generation. Fig. 6 presents an example of the nearest majority class instance selection among the majority class neighbors of \( \textbf{x}_{center} \). A disadvantage of the majority selection strategy is that the computational cost compared to the minority selection strategy may be higher, especially for datasets with high IR values.

\begin{figure}[H]
	\centering
	\includegraphics[width=12cm, keepaspectratio]{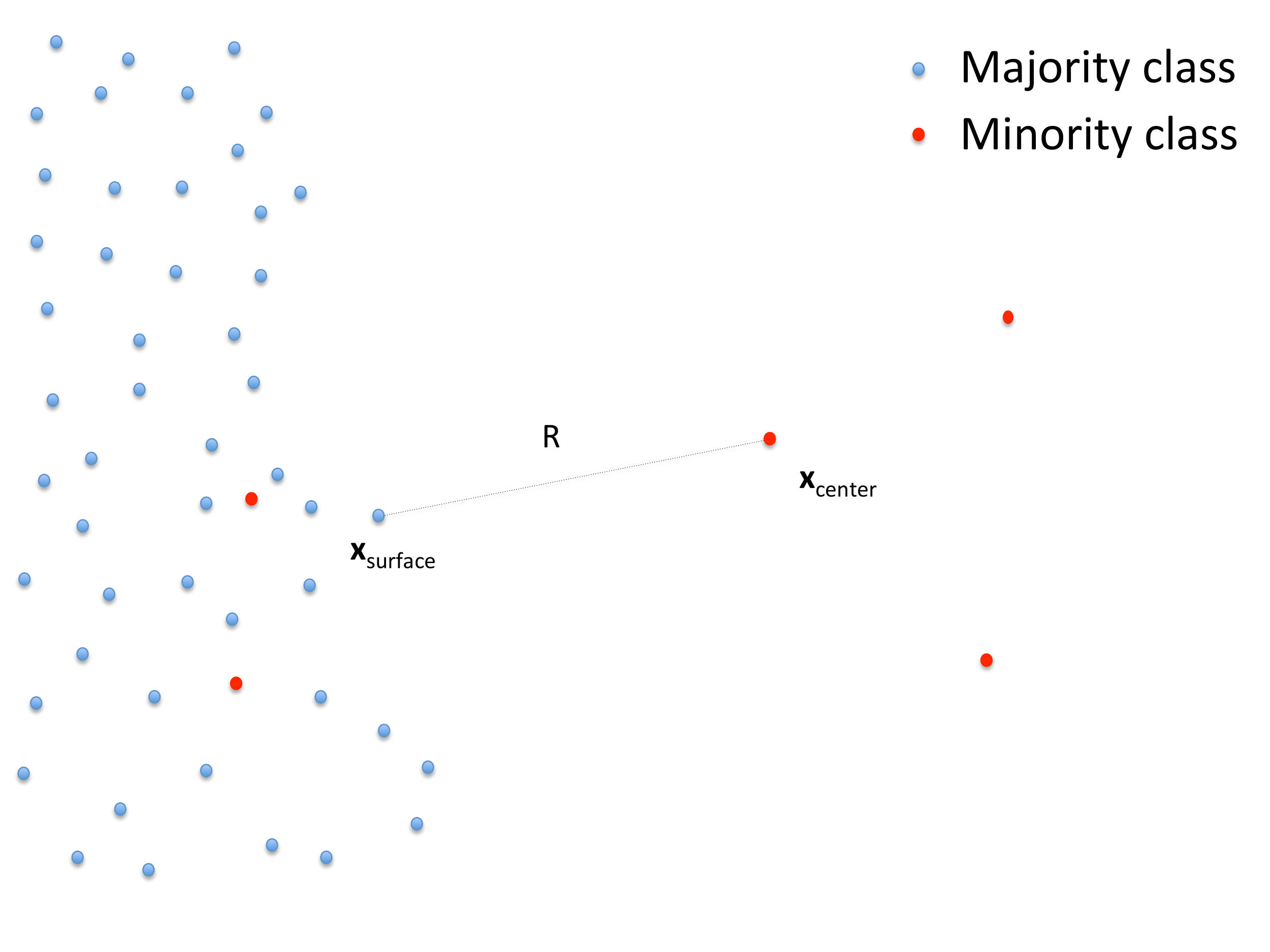}
	\captionbelow{An example of the majority selection strategy. A minority class instance is defined as the center of the hyper-spheroid and its closest majority class neighbor is selected as the surface point. Their distance is equal to the hyper-spheroid radius R.}
\end{figure}

2.4: The combined selection strategy initially applies the minority and majority selection strategies, identifying \( \textbf{x}_{min} \) and \( \textbf{x}_{maj} \) as the selected minority and majority class instances, respectively. The surface point \( \textbf{x}_{surface} \) is defined to be either \( \textbf{x}_{min} \) or \( \textbf{x}_{maj} \) so that its distance from the center \( \textbf{x}_{center} \) is minimized. This minimum distance is set equal to the radius R of the (truncated) hyper-spheroid with symmetry axis in the \( \textbf{x}_{surface} - \textbf{x}_{center} \) direction. Fig. 7 and fig.8 present both of these scenarios i.e. when \( \textbf{x}_{surface} \) is identified either as a minority or majority class instance. In any case since \( R \leq d_{maj} \) and the generation of artificial samples will occur in a distance \( d \) from \( \textbf{x}_{center} \) with \(d \leq R \), the above definitions imply that \( d \leq d_{maj} \). Therefore following the combined selection strategy, the expansion of the minority class area relative to the selected as a center minority class sample is restricted by the nearest majority class neighbor of the center, ensuring that the generation of noisy samples is avoided. Contrary to pure majority selection strategy, the expansion is not only safe but it is further restricted by the presence of minority class instances. The drawback of the combined, similarly to the majority selection strategy, is that it has a higher computational cost compared to the SMOTE/minority selection strategy.

\begin{figure}[H]
	\centering
	\includegraphics[width=12cm, keepaspectratio]{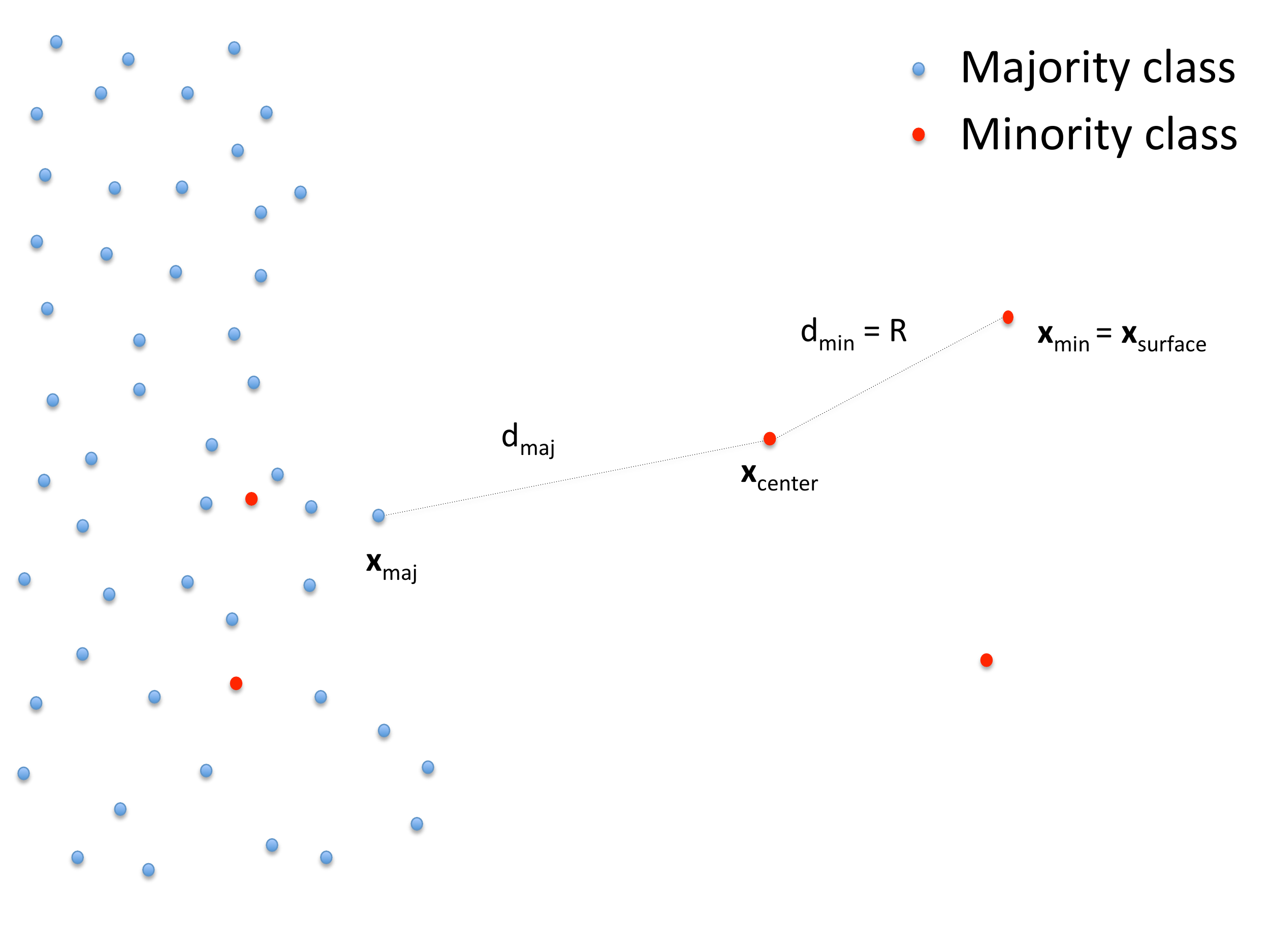}
	\captionbelow{A minority class sample is defined as the surface point since it is closer to the center than the nearest majority class instance.}
\end{figure}

\begin{figure}[H]
	\centering
	\includegraphics[width=12cm, keepaspectratio]{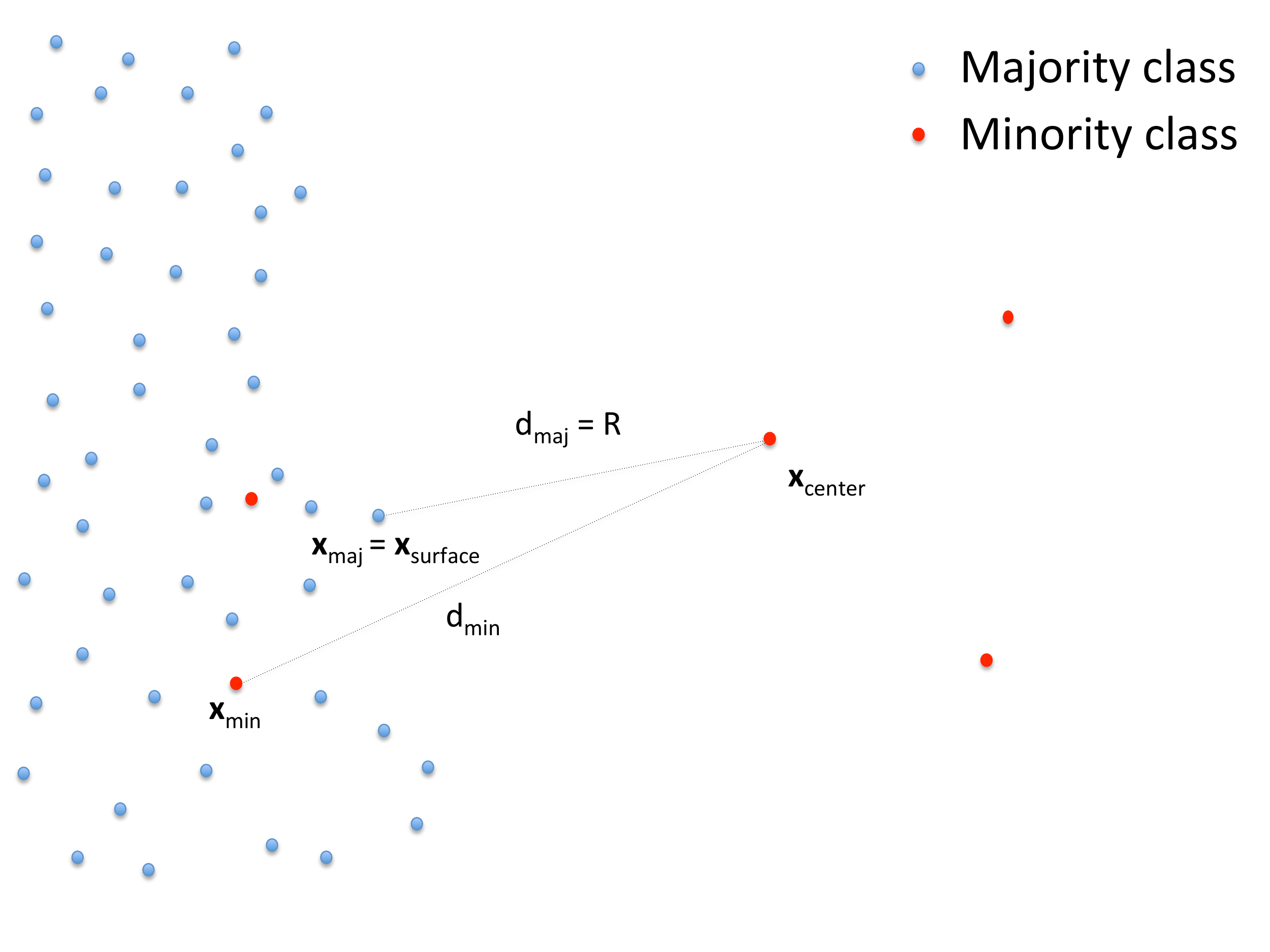}
	\captionbelow{The closest to the center majority class sample is defined as the surface point since it is closer to the center than the selected instance from the k nearest minority class neighbors of the center.}
\end{figure}

2.5: This step starts the data generation process. A random point \( \textbf{e}_{sphere} \) is generated on the surface of a unit hyper-sphere centered at the origin of the input space, using equation \eqref{eqn:sphere}. Applying equation \eqref{eqn:ball}, the point \( \textbf{e}_{sphere} \) is transformed to a random generated point \( \textbf{x}_{gen} \) inside the unit hyper-sphere. The final result of this process is a random generated point, uniformly distributed, within the unit hyper-sphere \cite{DasGupta2011}. Fig. 9 shows an example of this process in two dimensions.

\begin{figure}[H]
	\centering
	\includegraphics[width=12cm, keepaspectratio]{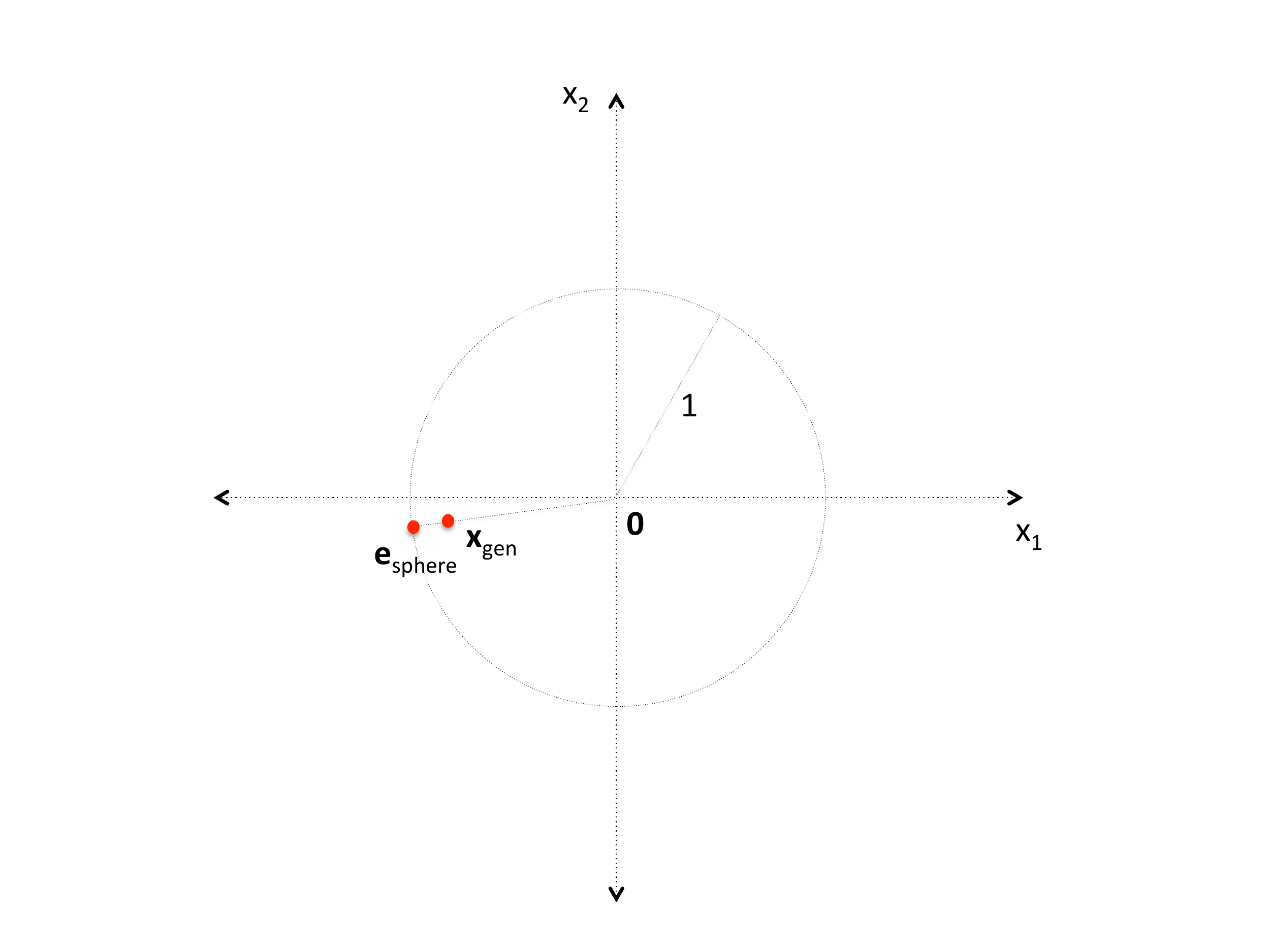}
	\captionbelow{A unit hyper-sphere centered at the origin of the input space. A point is randomly generated on the surface and moved to the interior of the unit hyper-sphere.}
\end{figure}

2.6: In this step a transformation is applied to the generated point \( \textbf{x}_{gen} \). We observe that the center \( \textbf{x}_{center} \) and the selected surface point \( \textbf{x}_{surface} \) define a special direction in the input space which is represented by the unit vector \( \textbf{e}_{\scriptscriptstyle//} \) of equation \eqref{eqn:unit}. SMOTE mechanism, in the case where the surface point is a minority class instance, exploits this direction by generating synthetic samples at the line segment between \( \textbf{x}_{center} \) and \( \textbf{x}_{surface} \). G-SMOTE algorithm parametrizes a generalized version of the SMOTE mechanism. More specifically, the unit vector \( \textbf{e}_{\scriptscriptstyle//} \) defines a family of parallel hyper-planes which are perpendicular to it. We define a linear mapping between \( \alpha_{trunc} \) and the point determined by the intersection of each hyper-plane and the parallel to \( \textbf{e}_{\scriptscriptstyle//} \) diameter. Therefore each one of these hyper-planes corresponds to a particular value of \( \alpha_{trunc} \) and partitions the hyper-sphere interior in to two areas. Let \( P \) the hyper-plane that passes through the origin and \( P' \) the hyper-plane for a specific non-zero value of \( \alpha_{trunc} \). When \( a_{trunc} > 0 \), the area that does not include the \( \textbf{e}_{\scriptscriptstyle//} \) point is truncated from the interior of the hyper-sphere, in the sense that if the \( \textbf{x}_{gen} \) point belongs to it then it is mapped with respect to \( P \) to the symmetric point \( \textbf{x}_{gen} - 2 \textbf{x}_{\scriptscriptstyle//} \) of equation  \eqref{eqn:truncate}, where \( \textbf{x}_{\scriptscriptstyle//} \) is defined in equation \eqref{eqn:parallel}. Condition  \eqref{eqn:condition} checks if  the \( \textbf{x}_{gen} \) is in the truncated area. Fig. 10 shows an example of the above transformation. When \( \alpha_{trunc} < 0 \), the transformation is similarly defined but in this case the truncation occurs in the area that includes the \( \textbf{e}_{\scriptscriptstyle//} \) point. In both cases, the absolute value of the hyper-parameter \( \alpha_{trunc} \) controls the extent of the truncation. Fig. 11 presents the truncated hyper-sphere areas for various positive and negative values of \( \alpha_{trunc} \). A final observation is that the above transformation effectively corresponds to a modification of the initial uniform probability distribution in the hyper-sphere. The truncated area acquires a zero value for the p.d.f., while its \( P \)-symmetric mapped area doubles its initial p.d.f. value.

\begin{figure}[H]
	\centering
	\includegraphics[width=12cm, keepaspectratio]{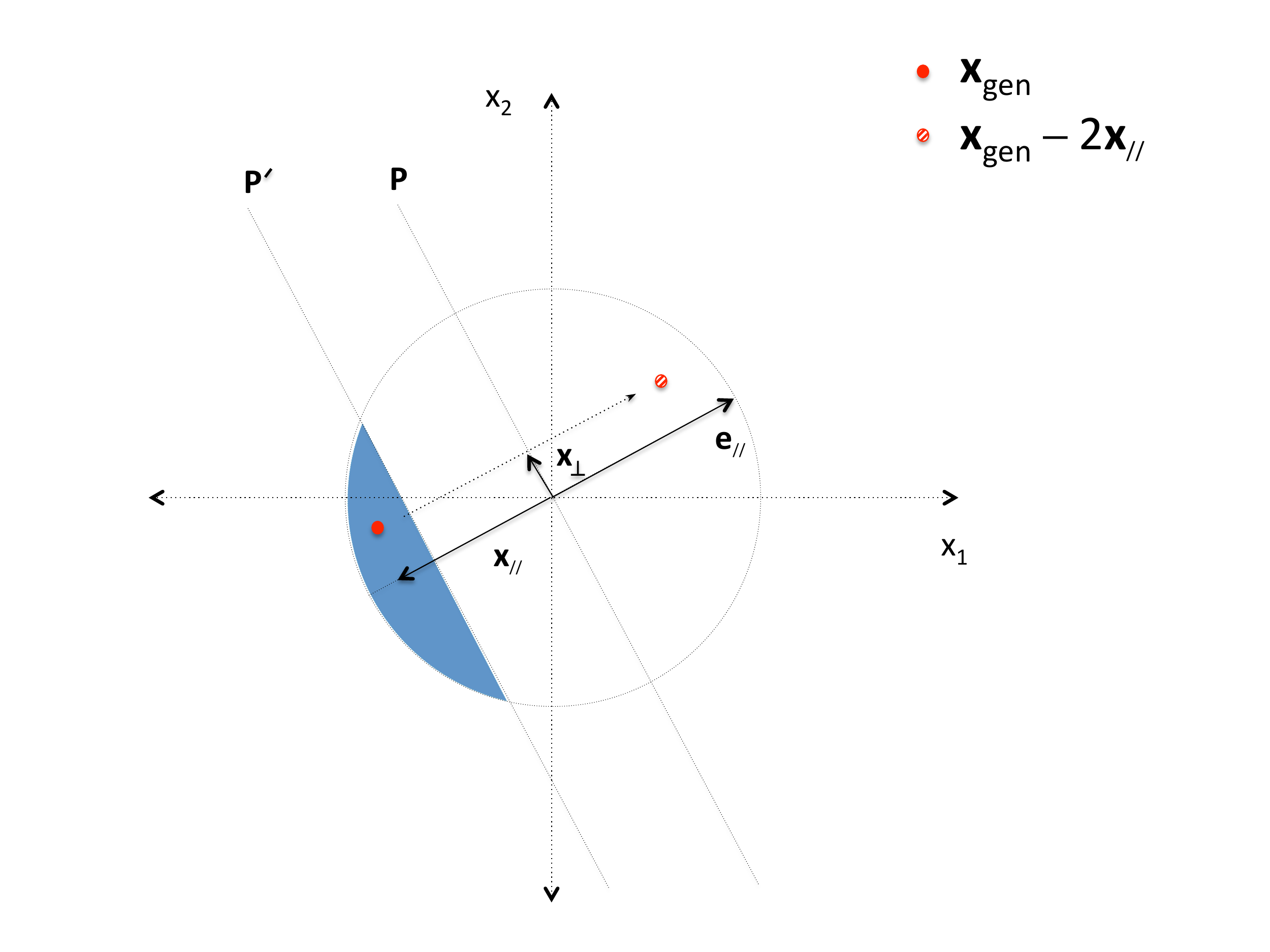}
	\captionbelow{An example of applying the \( \ truncate \) transformation. The shaded area corresponds to the resulting truncated area.}
\end{figure}

\begin{figure}[H]
	\centering
	\includegraphics[width=12cm, keepaspectratio]{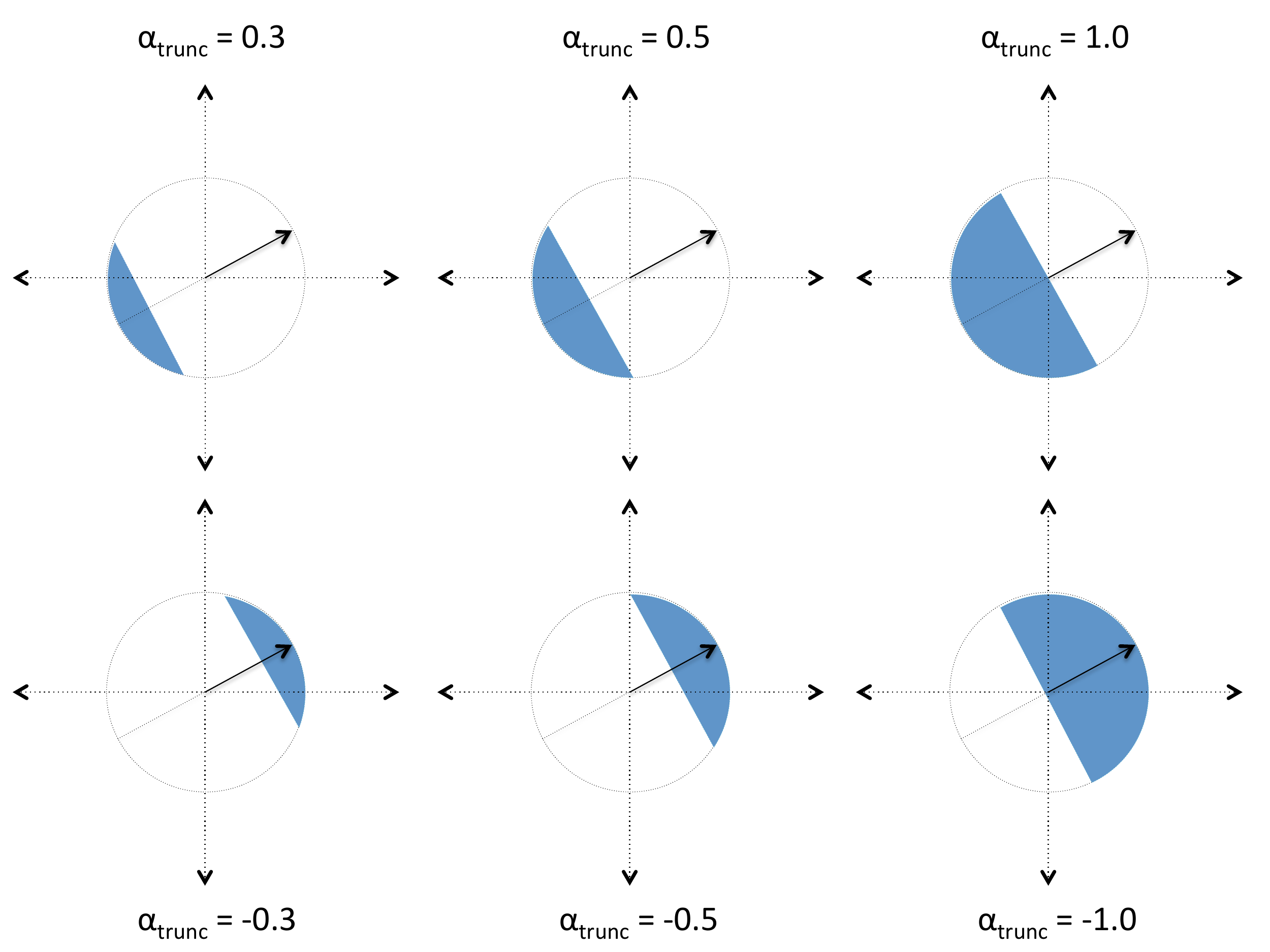}
	\captionbelow{Truncated areas for various values of \( \alpha_{trunc} \).}
\end{figure}

2.7: This step describes a transformation that corresponds to the deformation of the hyper-sphere in to a hyper-spheroid. More concretely, the point \( \textbf{x}_{gen} \) is moved to a perpendicular direction to the unit vector \( \textbf{e}_{\scriptscriptstyle//} \), towards the parallel to \( \textbf{e}_{\scriptscriptstyle//} \) diameter. This mapping is controlled by the \( \alpha_{def} \) hyper-parameter and from equation \eqref{eqn:deform} changes linearly with it. Therefore any point located at the surface of the hyper-sphere will remain to the surface of the new boundary while all the axes, except the one defined by the \( \textbf{e}_{\scriptscriptstyle//} \) unit vector, rescale by the factor \( \alpha_{def} \). This effectively corresponds to the formation of a hyper-spheroid boundary with symmetry axis at the \( \textbf{e}_{\scriptscriptstyle//} \) direction. Similarly to the truncation, the deformation transformation further modifies the initially uniform probability distribution. Fig. 12 presents a deformation of the unit hyper-sphere and the resulting mapping of the \( \textbf{x}_{gen} \) point. Fig. 13 shows the effect of increasing the \( \alpha_{def} \) values on the hyper-sphere deformation.

\begin{figure}[H]
	\centering
	\includegraphics[width=12cm, keepaspectratio]{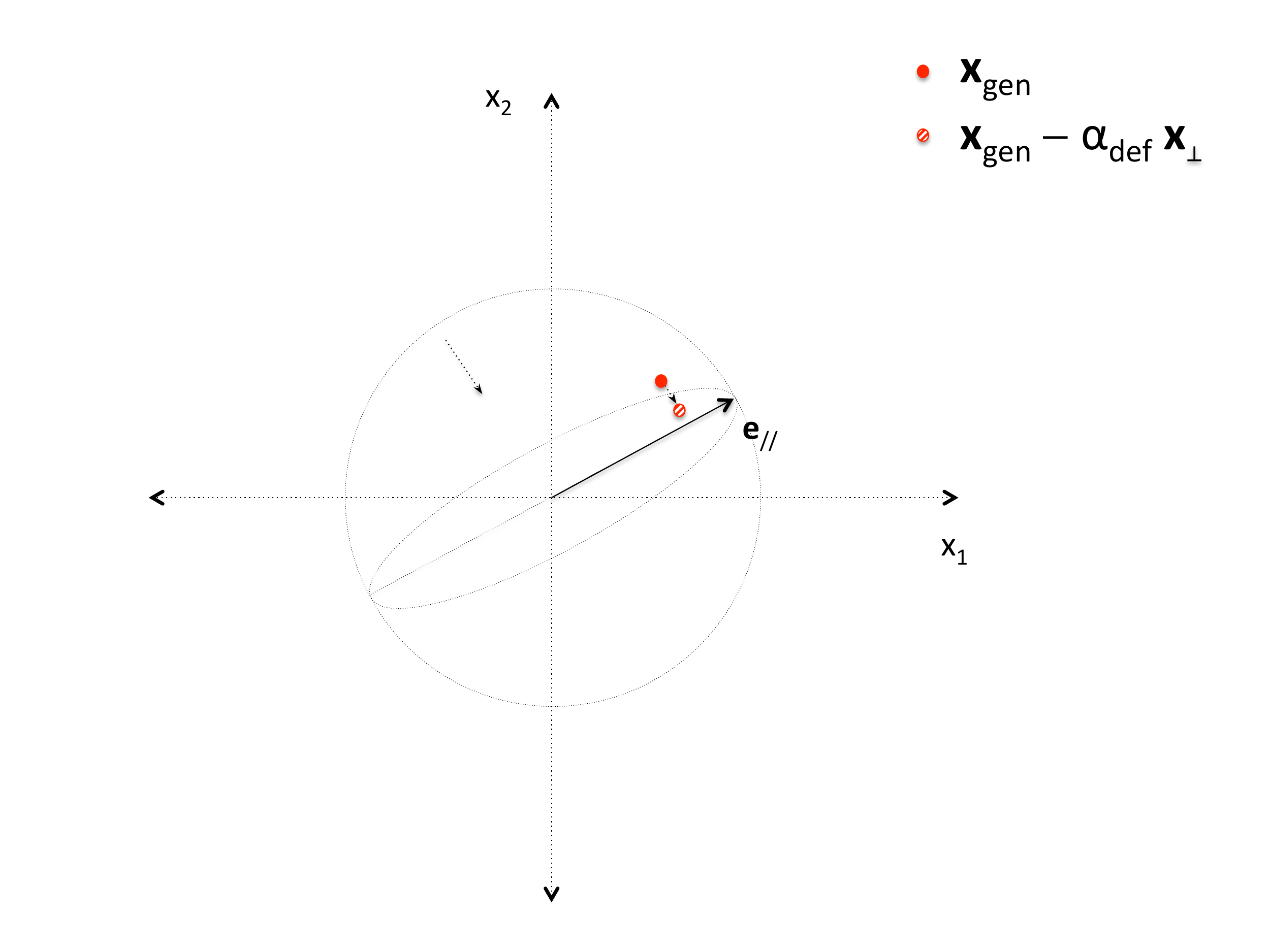}
	\captionbelow{A \( \ deform \) transformation is applied and the generated point is mapped to a new point towards the diameter of the hyper-sphere.}
\end{figure}

\begin{figure}[H]
	\centering
	\includegraphics[width=12cm, keepaspectratio]{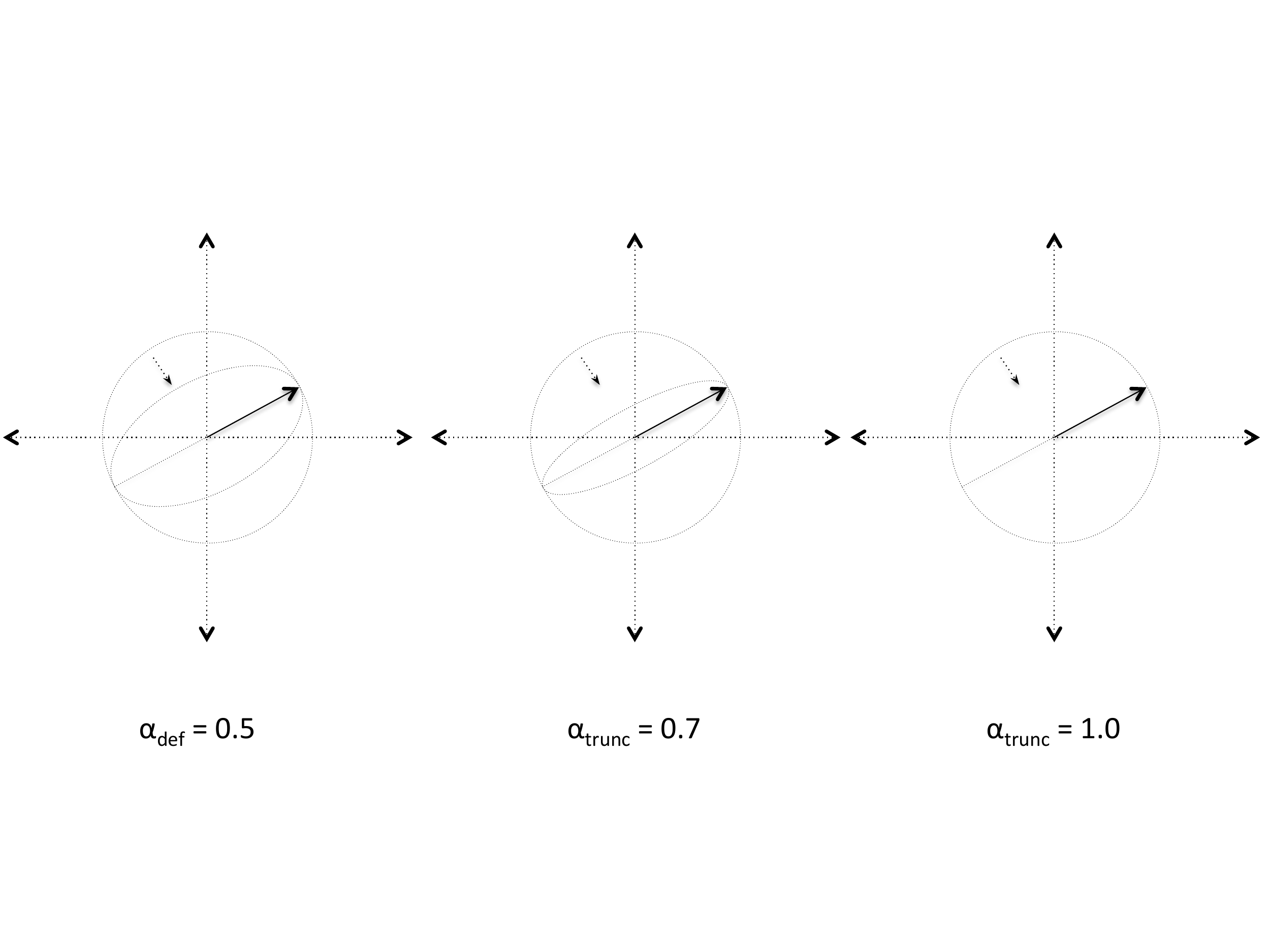}
	\captionbelow{The effect of increasing \( \alpha_{def} \) on the hypersphere deformation. The last case corresponds to a line segment.}
\end{figure}

2.8: The final step of the algorithm is the translation of the generated point by the \( \textbf{x}_{center} \) vector and the rescaling by the value of the radius \( R \). The combined result of this two transformations is described in equation \eqref{eqn:translate}. Fig. 14 and fig. 15 show the resulting boundaries of the permissible data generation area as well as a random generated point \( \textbf{x}_{center} \), of the two different scenarios presented in fig. 7 and fig. 8, after the application of truncation, deformation and translation.

\begin{figure}[H]
	\centering
	\includegraphics[width=12cm, keepaspectratio]{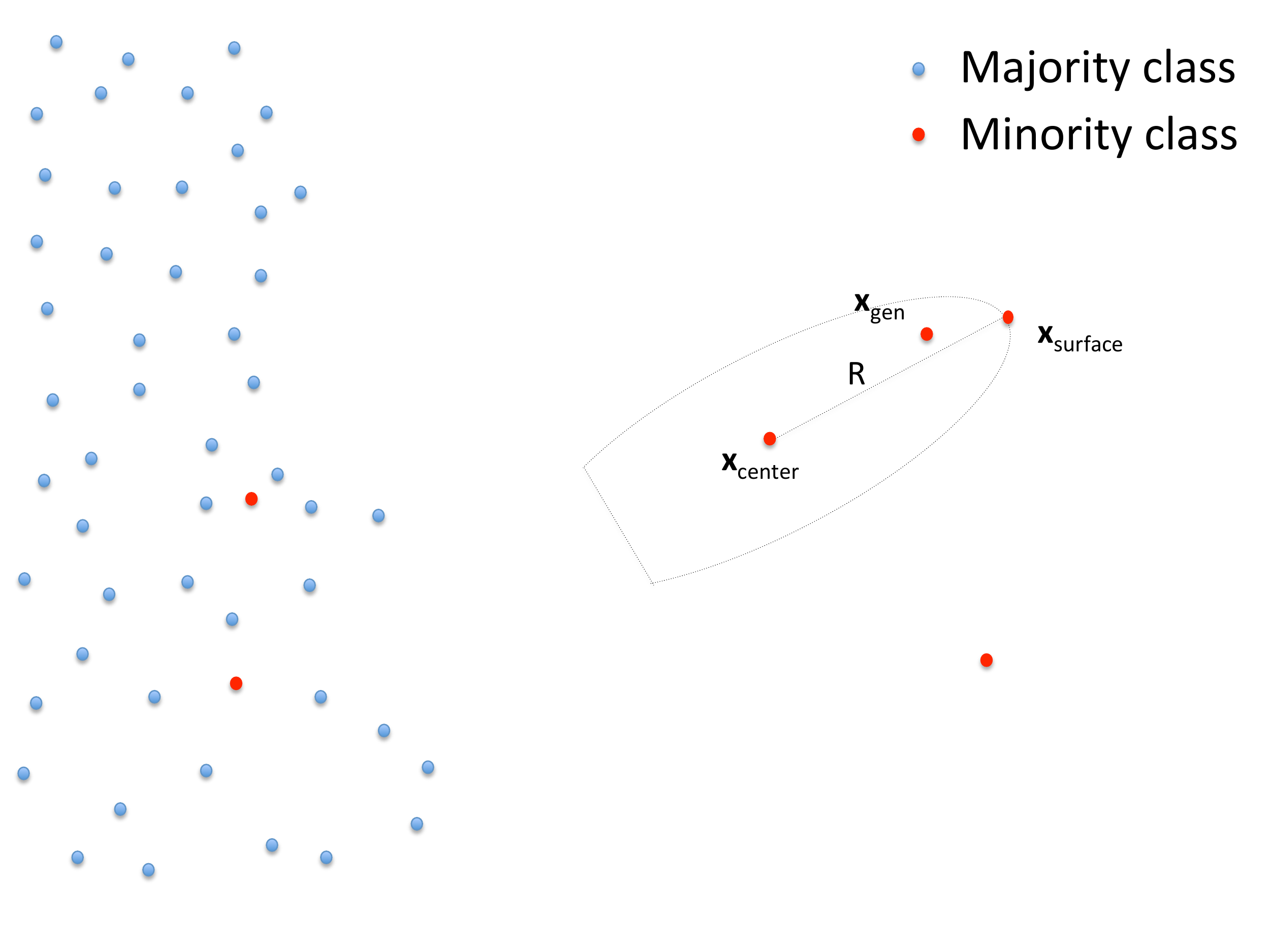}
	\captionbelow{Boundaries of permissible data generation area  for the scenario of fig 7.}
\end{figure}

\begin{figure}[H]
	\centering
	\includegraphics[width=12cm, keepaspectratio]{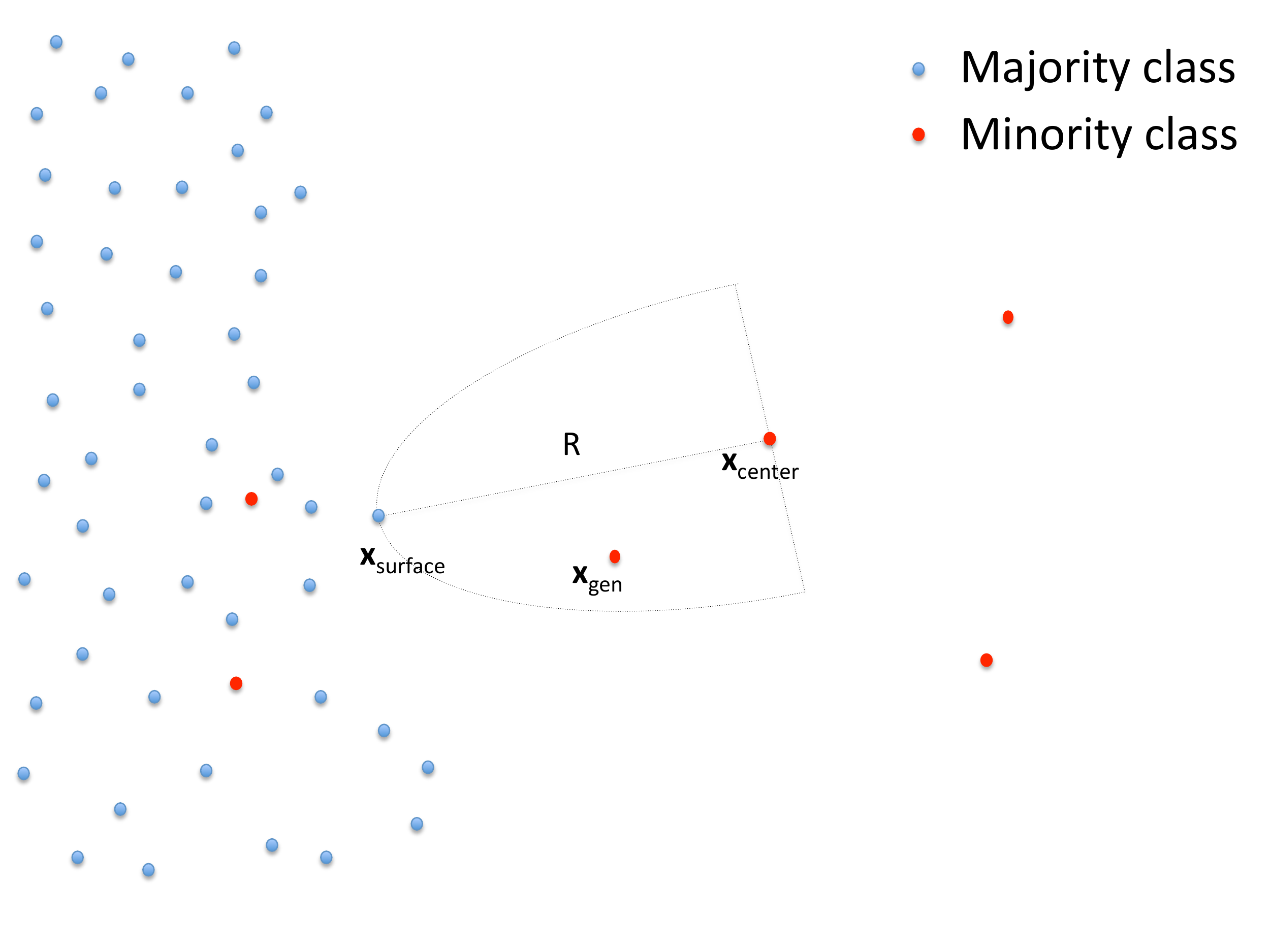}
	\captionbelow{Boundaries of permissible data generation area for the scenario of fig 8.}
\end{figure}

\section{Research methodology}

In order to test the performance of G-SMOTE we used 13 imbalanced data sets from the Machine Learning Repository UCI which have diverse characteristics. Table 1 shows a summary of these data sets:

\begin{table}[H]
	\centering
	\pgfplotstabletypesetfile[
		every head row/.style={before row=\toprule, after row=\midrule},
		every last row/.style={after row=\bottomrule},
		string type, col sep=comma]{summary.csv}
	\caption{Description of the datasets.}
\end{table}

The performance of G-SMOTE was evaluated and compared against the following oversampling methods:  SMOTE, Borderline SMOTE1, Borderline SMOTE2 and ADASYN. Since total accuracy is not appropriate for imbalanced datasets F-measure, G-mean and Area Under the ROC Curve (AUC) are used \cite{He2009}.

A ranking score was applied to each oversampling method for every combination of the 13 data sets, 3 evaluation metrics and 2 classifiers. Additionally to the 5 oversampling algorithms we also included the performance of the classifiers when no oversampling is used. Therefore the ranking score for the best performing method is 1 and for the worst performing method is 6. The Friedman test was applied to the ranking results. Generally the Friedman test is used to detect differences in the results across multiple experimental attempts, when the normality assumption may not hold. The null hypothesis is whether the classifiers have a similar performance across the oversampling methods and evaluation metrics when they are compared to their mean rankings.

For the evaluation of the oversampling methods, Logistic Regression (LR) \cite{McCullagh1989} and Gradient Boosting Classifier (GBC) \cite{Friedman2001} were used. In order to evaluate the performance of the algorithms \( n \)-fold cross validation was applied with \( n = 5 \). Before training, in each stage \(i \in \{1, 2 ,... , n \} \) of the \( n \)-fold cross validation procedure, synthetic data \( T_{g, i} \) were generated based on the training data \(T_{i} \) of the \( n - 1 \) folds such that the resulting \(T_{g, i} \cup T_{i} \) training set becomes perfectly balanced. This enhanced training set in turn was used to train the classifier. The performance evaluation of the classifiers was done on the validation data \( V_{i} \) of the remaining fold. 

A variety of hyper-parameters were used for the classifiers and the over-samplers, and the highest cross validation score for each combination of datasets, classifiers, over-samplers and evaluation metrics was reported. More specifically, the GBC hyper-parameter grid included the four combinations resulting from  \textit{max depth} \( \in \{ 5 , 8 \} \) and \textit{number of estimators} \( \in \{50, 100\} \). For SMOTE and the two kinds of Borderline SMOTE, the optimal value of \( k \) nearest neighbors was selected as \( k \in \{ 3, 4 \} \), while for ADASYN the same value was set to \( k = 3 \). Finally a hyper-parameter grid was generated for G-SMOTE including the three different selection strategies, the number of nearest neighbors \(k \in \{ 3, 4 \} \), the truncation factor \( \alpha_{trunc} \in \{ -1.0, 0.0, 0.5, 1.0 \} \) and the deformation factor \( \alpha_{def} \in \{ 0.0, 0.5, 1.0 \} \).

The experimental procedure was repeated 5 times and the reported results include the average values between the experiments. The implementation of the classifiers and standard oversampling algorithms was based on the Python libraries Scikit-Learn \cite{Pedregosa2011} and Imbalanced-Learn \cite{Lemaitre2016}.

\section{Experimental results}

The cross validation results for each combination of datasets, classifiers, over-samplers and metrics are shown in Table 2:

\pgfplotstableset{
	begin table=\begin{longtabu},
	end table=\end{longtabu},
}

\begin{filecontents*}{cv.csv}
Dataset,Classifier,Metric,No oversampling,SMOTE,Borderline SMOTE1,Borderline SMOTE2,ADASYN,Geometric SMOTE
Breast,GBC,F,0.700,0.715,0.727,0.705,0.673,0.722
Breast,GBC,G,0.770,0.780,0.795,0.774,0.748,0.789
Breast,GBC,AUC,0.867,0.867,0.867,0.869,0.858,0.874
Breast,LR,F,0.687,0.732,0.747,0.734,0.730,0.748
Breast,LR,G,0.752,0.799,0.812,0.803,0.798,0.812
Breast,LR,AUC,0.884,0.890,0.888,0.889,0.889,0.896
Ecoli,GBC,F,0.781,0.777,0.773,0.770,0.662,0.821
Ecoli,GBC,G,0.868,0.872,0.868,0.862,0.861,0.904
Ecoli,GBC,AUC,0.948,0.945,0.945,0.940,0.937,0.960
Ecoli,LR,F,0.249,0.716,0.682,0.641,0.473,0.718
Ecoli,LR,G,0.378,0.894,0.885,0.870,0.759,0.900
Ecoli,LR,AUC,0.934,0.935,0.929,0.924,0.896,0.936
Eucalyptus,GBC,F,0.551,0.526,0.540,0.535,0.506,0.549
Eucalyptus,GBC,G,0.682,0.699,0.703,0.714,0.688,0.730
Eucalyptus,GBC,AUC,0.878,0.869,0.872,0.87,0.841,0.874
Eucalyptus,LR,F,0.207,0.529,0.530,0.513,0.468,0.540
Eucalyptus,LR,G,0.399,0.790,0.787,0.785,0.757,0.797
Eucalyptus,LR,AUC,0.857,0.886,0.883,0.879,0.870,0.890
Glass,GBC,F,0.779,0.774,0.774,0.774,0.758,0.792
Glass,GBC,G,0.831,0.830,0.833,0.836,0.822,0.846
Glass,GBC,AUC,0.924,0.920,0.918,0.919,0.911,0.929
Glass,LR,F,0.505,0.657,0.648,0.651,0.650,0.661
Glass,LR,G,0.613,0.733,0.715,0.714,0.719,0.734
Glass,LR,AUC,0.824,0.825,0.814,0.818,0.819,0.825
Haberman,GBC,F,0.339,0.393,0.388,0.385,0.376,0.426
Haberman,GBC,G,0.505,0.558,0.551,0.549,0.542,0.584
Haberman,GBC,AUC,0.627,0.637,0.641,0.636,0.619,0.668
Haberman,LR,F,0.239,0.477,0.485,0.481,0.418,0.484
Haberman,LR,G,0.376,0.622,0.632,0.628,0.578,0.628
Haberman,LR,AUC,0.686,0.693,0.688,0.684,0.650,0.694
Heart,GBC,F,0.745,0.750,0.753,0.754,0.739,0.761
Heart,GBC,G,0.771,0.775,0.776,0.779,0.765,0.781
Heart,GBC,AUC,0.859,0.862,0.864,0.864,0.856,0.868
Heart,LR,F,0.822,0.822,0.823,0.824,0.820,0.826
Heart,LR,G,0.840,0.840,0.841,0.841,0.837,0.843
Heart,LR,AUC,0.902,0.903,0.901,0.900,0.901,0.903
Iris,GBC,F,0.923,0.937,0.915,0.922,0.911,0.939
Iris,GBC,G,0.944,0.954,0.939,0.945,0.940,0.956
Iris,GBC,AUC,0.980,0.985,0.978,0.984,0.972,0.988
Iris,LR,F,0.334,0.648,0.637,0.636,0.674,0.657
Iris,LR,G,0.453,0.731,0.715,0.712,0.752,0.739
Iris,LR,AUC,0.793,0.792,0.747,0.745,0.803,0.801
Libra,GBC,F,0.780,0.850,0.839,0.876,0.786,0.924
Libra,GBC,G,0.821,0.890,0.875,0.915,0.847,0.946
Libra,GBC,AUC,0.943,0.969,0.962,0.976,0.931,0.987
Libra,LR,F,0.316,0.575,0.512,0.475,0.631,0.565
Libra,LR,G,0.432,0.745,0.686,0.659,0.779,0.738
Libra,LR,AUC,0.737,0.766,0.733,0.708,0.776,0.765
Liver,GBC,F,0.636,0.630,0.639,0.644,0.642,0.657
Liver,GBC,G,0.688,0.681,0.687,0.693,0.686,0.701
Liver,GBC,AUC,0.753,0.752,0.755,0.754,0.756,0.758
Liver,LR,F,0.579,0.633,0.622,0.629,0.626,0.640
Liver,LR,G,0.644,0.669,0.658,0.661,0.650,0.669
Liver,LR,AUC,0.713,0.712,0.707,0.709,0.710,0.715
Pima,GBC,F,0.623,0.651,0.654,0.652,0.644,0.659
Pima,GBC,G,0.704,0.728,0.731,0.729,0.722,0.734
Pima,GBC,AUC,0.812,0.811,0.806,0.806,0.800,0.822
Pima,LR,F,0.617,0.675,0.675,0.674,0.676,0.677
Pima,LR,G,0.692,0.748,0.747,0.746,0.747,0.749
Pima,LR,AUC,0.825,0.827,0.825,0.824,0.824,0.830
Segment,GBC,F,0.925,0.927,0.912,0.887,0.856,0.934
Segment,GBC,G,0.941,0.967,0.961,0.960,0.956,0.970
Segment,GBC,AUC,0.996,0.997,0.995,0.994,0.990,0.997
Segment,LR,F,0.648,0.645,0.634,0.622,0.552,0.640
Segment,LR,G,0.749,0.881,0.881,0.888,0.850,0.881
Segment,LR,AUC,0.942,0.942,0.930,0.923,0.922,0.944
Vehicle,GBC,F,0.923,0.927,0.929,0.925,0.923,0.935
Vehicle,GBC,G,0.951,0.958,0.958,0.961,0.962,0.969
Vehicle,GBC,AUC,0.994,0.994,0.995,0.994,0.993,0.995
Vehicle,LR,F,0.940,0.941,0.940,0.895,0.816,0.940
Vehicle,LR,G,0.961,0.968,0.967,0.958,0.924,0.969
Vehicle,LR,AUC,0.995,0.995,0.995,0.992,0.983,0.995
Wine,GBC,F,0.911,0.908,0.912,0.907,0.875,0.933
Wine,GBC,G,0.925,0.923,0.927,0.922,0.898,0.943
Wine,GBC,AUC,0.979,0.978,0.978,0.975,0.965,0.986
Wine,LR,F,0.928,0.935,0.939,0.937,0.898,0.938
Wine,LR,G,0.941,0.947,0.951,0.950,0.918,0.951
Wine,LR,AUC,0.992,0.992,0.991,0.992,0.985,0.993
\end{filecontents*}
\pgfplotstabletypeset
[
font={\small},
begin table=\begin{longtabu} to \linewidth {@{}ccc*6{X[c]}@{}},
end table=\end{longtabu},
skip coltypes=true,
every head row/.style={before row=\toprule, after row=\midrule\endhead},
every last row/.style={after row=\bottomrule \\\caption{Cross validation score results for all combinations of datasets, classifiers, metrics and over-samplers.}},
string type, col sep=comma
]
{cv.csv}

As explained in section 5, for every row of table 2, a ranking score in the range 1 to 6 is assigned to each over-sampler. Then the mean ranking of the oversampling methods across the data sets for each combination of a classifier and evaluation metric is presented in Table 3:

\begin{filecontents*}{mean_ranking.csv}
Classifier,Metric,No oversampling,SMOTE,Borderline SMOTE1,Borderline SMOTE2,ADASYN,Geometric SMOTE
GBC,F,4.08,3.38,3.23,3.69,5.46,1.15
GBC,G,4.85,3.46,3.31,3.15,5.15,1.08
GBC,AUC,3.38,3.46,3.54,3.77,5.69,1.15
LR,F,5.08,2.92,3.15,3.85,4.31,1.69
LR,G,5.62,2.85,3.00,3.62,4.31,1.62
LR,AUC,3.46,2.38,4.54,4.92,4.46,1.23
\end{filecontents*}
\pgfplotstabletypeset
[
font={\small},
begin table=\begin{longtabu} to \linewidth {@{}ccc*5{X[c]}@{}},
	end table=\end{longtabu},
skip coltypes=true,
every head row/.style={before row=\toprule, after row=\midrule\endhead},
every last row/.style={after row=\bottomrule \\\caption{Results for mean ranking of over-samplers across the datasets.}},
string type, col sep=comma
]
{mean_ranking.csv}

The results of the application of the Friedman test are shown in Table 4:

\pgfplotstabletypesetfile[
every head row/.style={before row=\toprule, after row=\midrule},
every last row/.style={after row=\bottomrule \\\caption{Results for Friedman's test.}},
string type, col sep=comma]{friedman.csv}

Therefore at a significance level of a = 0.05 the null hypothesis is rejected, i.e. the classifiers do not perform similarly in the mean rankings across the oversampling methods and evaluation metrics. Additionally from the mean ranking results of table 3 we conclude that G-SMOTE performs better than the other oversampling methods.

\section{Conclusions}

In this paper we presented G-SMOTE, a new oversampling algorithm, that extends the SMOTE data generation mechanism. G-SMOTE selects a safe radius around each minority class instance and generates artificial data within a (truncated) hyper-spheroid. G-SMOTE performance was evaluated on 13 datasets with different imbalance ratios and compared to multiple oversampling methods, using Logistic Regression and Gradient Boosting Machine as classifiers. 

The results show that G-SMOTE performs better compared to the other methods.The explanation for this improvement in performance relates to the ability of G-SMOTE to generate a variety of artificial data in safe areas of the input space, while, at the same time, aggressively increasing the diversity of the minority class. G-SMOTE parametrizes efficiently the data generation process and adapts to the special characteristics of each imbalanced dataset.

G-SMOTE can be a useful tool for researchers and practitioners since it results in the generation of high quality artificial data and only requires the tuning of a small number of parameters.

\bibliography{references}
\bibliographystyle{apalike}

\end{document}